\documentclass[nohyperref]{article}
\usepackage{microtype}
\usepackage{graphicx}
\usepackage{subfigure}

\usepackage{booktabs}

\usepackage{bm}
\usepackage{nicefrac}
\usepackage{adjustbox}

\usepackage{hyperref}

\usepackage[accepted]{icml2022}

\usepackage{amsmath}
\usepackage{amssymb}
\usepackage{mathtools}
\usepackage{amsthm}

\usepackage[capitalize,noabbrev]{cleveref}

\theoremstyle{plain}

\theoremstyle{definition}

\theoremstyle{remark}

\graphicspath{{./Figures/}}

\renewcommand{\vec}[1]{\bm{#1}}
\newcommand{\mat}[1]{\bm{#1}}

\newcommand{\set}[1]{\mathcal{#1}}

\newcommand{\trn}[1]{#1^\intercal}

\newcommand{\ipt}[2]{\trn{#1} #2}
\newcommand{\opt}[2]{#1 \trn{#2}}

\newcommand{\st}{\operatorname{s.\!t.}}

\newcommand{\amin}[1]{\operatorname*{argmin}_{#1}}

\newcommand{\ket}[1]{\vert {#1} \rangle}

\newcommand{\bra}[1]{\langle {#1} \vert}

\theoremstyle{definition}
\newtheorem*{defn}{Definition}

\definecolor{ml2rblu}{rgb}{0.02,0.27,0.45}
\definecolor{ml2ryel}{rgb}{0.98,0.72,0.18}
\definecolor{ml2rgrn}{rgb}{0.50,0.71,0.18}
\definecolor{ml2rtrq}{rgb}{0.00,0.57,0.57}

\icmltitlerunning{Gradient Flows for $L_2$ Support Vector Machine Training}

\begin{document}

\onecolumn

\begin{icmlauthorlist}
\icmltitle{Gradient Flows for $L_2$ Support Vector Machine Training}
\icmlauthor{Christian Bauckhage}{unib,iais}
\icmlauthor{Helen Schneider}{iais}
\icmlauthor{Benjamin Wulff}{iais}
\icmlauthor{Rafet Sifa}{iais}
\end{icmlauthorlist}

\icmlaffiliation{unib}{University of Bonn, Germany}
\icmlaffiliation{iais}{Fraunhofer IAIS, Germany}
\icmlcorrespondingauthor{Christian Bauckhage}{first.last@iais.fraunhofer.de}
\vskip 0.3in

%\twocolumn[
%\icmltitle{Training SVMs by Solving Differential Equations}
%
%\icmlsetsymbol{equal}{*}
%
%\begin{icmlauthorlist}
%\icmlauthor{Firstname1 Lastname1}{equal,yyy}
%\icmlauthor{Firstname2 Lastname2}{equal,yyy,comp}
%\icmlauthor{Firstname3 Lastname3}{comp}
%\icmlauthor{Firstname4 Lastname4}{sch}
%\end{icmlauthorlist}
%
%\icmlaffiliation{yyy}{Department of XXX, University of YYY, Location, Country}
%\icmlaffiliation{comp}{Company Name, Location, Country}
%\icmlaffiliation{sch}{School of ZZZ, Institute of WWW, Location, Country}
%
%\icmlcorrespondingauthor{Firstname1 Lastname1}{first1.last1@xxx.edu}
%\icmlcorrespondingauthor{Firstname2 Lastname2}{first2.last2@www.uk}
%
%% You may provide any keywords that you
%% find helpful for describing your paper; these are used to populate
%% the "keywords" metadata in the PDF but will not be shown in the document
%\icmlkeywords{Machine Learning, ICML}
%
%\vskip 0.3in
%]

%\printAffiliationsAndNotice{\icmlEqualContribution}
\printAffiliationsAndNotice{}

\begin{abstract}
We explore the merits of training of support vector machines for binary classification by means of solving systems of ordinary differential equations. We thus assume a continuous time perspective on a machine learning problem which may be of interest for implementations on (re)emerging hardware platforms such as analog- or quantum computers.
\end{abstract}

\section{Introduction}

In this short paper, we consider the idea of training support vector machines (SVMs) for binary classification by means of solving systems of ordinary differential equations. To this end, we devise continuous time gradient flows over the feasible set of a corresponding dual training problem. These flows are known to converge to asymptotically stable stationary points from which one can compute the parameters of the sought after classifier.

Against the backdrop of several recent developments, our motivations for using continuous time models for SVM training are at least fourfold: 

1) There is renewed interest in SVMs because connections to deep neural networks have recently been worked out \cite{Jacot2018-NTK,Arora2019-OEC,Chen2021-OTE}. Gradient flows are an important tool to establish this connection and, as the flow we consider below differs from previously considered ones, it might inform further analysis.

2) Systems of ordinary differential equations (ODEs) occur in all of the hard sciences. Often, they do not have closed form solutions but there exist numerous methods and software packages for numerical ODE solving. Hence, by setting up SVM training as a continuous time problem, one gains access to domain-agnostic learning algorithms and versatile computational paradigms that can be implemented on various kinds of hardware platforms.

3) There are growing concerns as to the environmental sustainability of (deep) learning on GPU clusters \cite{Anthony2020-CTA,Thompson2021-DLD}. This has prompted researchers to (re)consider energy efficient analog computing \cite{Schuman2017-ASO,Haensch2019-TNG}. Indeed, analog circuits composed of resistors, capacitors, inductors, and operational amplifiers allow for differential equation solving \cite{Ulmann2013-AC}. This, in turn, suggests that SVMs should be trainable on low cost, low power hardware.

4) There also is increasing interest in quantum machine learning. Crucially, the evolution of the state of a quantum computer is governed by the Schr\"odinger equation, i.e.~by a differential equation. These dynamics can be used for differential equation solving \cite{Lloyd2020-QAF,Liu2021-EQA,Zanger2021-QAF} which, in turn, suggests that a continuous time perspective on SVM training could lead to novel quantum learning systems.  
%\cite{Srivastava2019-BAF}

In what follows, we will particularly focus on $L_2$ SVMs. This choice is admittedly informed by personal preferences. However, it does not constitute a loss of generality but ---seen from the point of view of mathematical intricacy--- marks a middle ground: gradient flows for $L_2$ SVM training are easier to set up than those for conventional SVMs and not as trivial as those for least squares SVMs. 

As $L_2$ SVMs are arguably lesser known than other SVMs, we next briefly summarize the underlying concepts. In section~\ref{sec:flows}, we then discuss training via solving gradient flows and, in section~\ref{sec:conclusion}, we conclude with an outlook to auspicious future work.

\section{Preliminaries}

This section ever so briefly recalls the theory behind $L_2$ SVMs; readers familiar with this topic may safely skip ahead.

Since we address binary classifier training, we assume that we are given labeled training data $\bigl\{ ( \vec{x}_i, y_i ) \bigr\}_{i=1}^n$ where the data $\vec{x}_i \in \mathbb{R}^m$ have been sampled from two classes and the labels $y_i \in \{-1, +1\}$ indicate class membership. To be able to write compact mathematical expressions, we gather the data points in an $m \times n$ matrix and their labels in an $n$ vector, namely $\mat{X} = \bigl[ \vec{x}_1, \vec{x}_2, \ldots, \vec{x}_n \bigr]$ and $\vec{y} = \trn{\bigl[ \,y_1, \,y_2, \ldots, \,y_n \bigr]}$.

Training a classifier for such data is to estimate the parameters of a function $y : \mathbb{R}^m \rightarrow \{-1, +1\}$. A generic ansatz is a linear classifier $y (\vec{x}) = \operatorname*{sign} \bigl( \ipt{\vec{x}}{\vec{w}} - \theta \bigr)$ with weight vector $\vec{w} \in \mathbb{R}^m$ and a threshold value $\theta \in \mathbb{R}$. What turns $y (\vec{x})$ into an SVM is the idea of estimating its weights such that projection $\ipt{\vec{x}_i}{\vec{w}}$ of the training data from both classes are maximally separated. As there exist various loss functions for max-margin criteria, SVMs come in different flavors. Well known are those of Cortes and Vapnik \yrcite{Cortes1995-SVN} or least squares SVMs due to Suykens and Venderwalle\yrcite{Suykens1999-LSS}. Another variant are $L_2$ SVMs dating back to work by Frie{\ss} and Harrison \yrcite{Friess1998-TKA} and Mangasarian and Musicant \yrcite{Mangasarian2001-LSVM}. These are of practical interest as they are easy to train \cite{Wu2010-TGT,Alaiz2018-MFW,Sifa2018-SRN} and we next sketch how.

\subsection{$L_2$ Support Vector Machines}

A (linear) support vector machine for binary classification determines the max-margin hyperplane between the training data for two given classes. Should these not be linearly separable, one typically incorporates slack variables whose influence is controlled using a parameter $C \geq 0 \in \mathbb{R}$. When training an $L_2$ SVM, slack variables enter the primal objective in form of a sum of squares which differs from conventional SVMs. Similar to conventional SVMs, the primal problem of training an $L_2$ SVM comes with inequality constraints which differs from least squares SVMs. Evaluating the Karush-Kuhn-Tucker conditions of optimality \cite{Bauckhage2021-SVMdual}, reveals the \emph{dual problem of training an $L_2$ SVM} 
\begin{gather}
\label{eq:L2dual}
\begin{aligned}
  \amin{\vec{\mu}} & \;\; \tfrac{1}{2} \, \trn{\vec{\mu}} \Bigl[ \ipt{\mat{X}}{\mat{X}} \odot \opt{\vec{y}}{\vec{y}} + \opt{\vec{y}}{\vec{y}} + \tfrac{1}{C} \, \mat{I} \Bigr] \, \vec{\mu} \\[1ex]
  \st &  
  \begin{aligned} 
  \;\; \ipt{\vec{1}}{\vec{\mu}} & = 1 \\[1ex]
  \;\; \vec{\mu} & \succeq \vec{0} 
  \end{aligned}
\end{aligned}
\end{gather}
Here, $\mat{I}$ denotes the $n \times n$ identity matrix, $\vec{0}, \vec{1} \in \mathbb{R}^n$ are the vectors of all zeros and ones, $\vec{\mu} \in \mathbb{R}^n$ is a vector of $n$ Lagrange multipliers $\mu_i$, and $\odot$ denotes the Hadamard product, i.e.~the element-wise product of vectors or matrices. Once the minimizer of \eqref{eq:L2dual} has been found, those elements $\mu_s$ of $\vec{\mu}$ which exceed zero identify which training data points support the sought after hyperplane. This, in turn, allows for computing $\vec{w} = \mat{X} \bigl[ \vec{y} \odot \vec{\mu} \bigr]$ and $\theta = - \trn{\vec{1}} \bigl[ \vec{y} \odot \vec{\mu} \bigr]$ so that the sought after classifier becomes $y(\vec{x}) = \operatorname*{sign} \bigl( \ipt{\vec{x}}{\mat{X} \bigl[ \vec{y} \odot \vec{\mu} \bigr]} + \trn{\vec{1}} \bigl[ \vec{y} \odot \vec{\mu} \bigr]  \bigr) = \operatorname*{sign} \bigl( \bigl[ \ipt{\vec{x}}{\mat{X}} + \trn{\vec{1}} \bigr] \bigl[ \vec{y} \odot \vec{\mu} \bigr] \bigr)$.

\subsection{Invoking the Kernel Trick}

In training and application of an $L_2$ SVM, data vectors exclusively occur within inner products, namely $\ipt{\mat{X}}{\mat{X}}$ and $\ipt{\vec{x}}{\mat{X}}$, respectively. This allows for invoking the kernel trick and thus for treating non-linear settings. In other words, considering a Mercer kernel $k : \mathbb{R}^m \times \mathbb{R}^m \rightarrow \mathbb{R}$, a non-linear classifier can be trained by replacing the Gram matrix $\ipt{\mat{X}}{\mat{X}}$ with a kernel matrix $\mat{K}$ whose elements are given by $K_{ij} = k \bigl( \vec{x}_i, \vec{x}_j \bigr)$.  By the same token, the trained classifier then becomes $y(\vec{x}) = \operatorname*{sign} \bigl( \bigl[ \trn{\vec{k}}(\vec{x}) + \trn{\vec{1}} \bigr] \bigl[ \vec{y} \odot \vec{\mu} \bigr] \bigr)$ where the elements of the kernel vector $\vec{k}(\vec{x})$ amount to $k_i(\vec{x}) = k \bigl( \vec{x}, \vec{x}_i \bigr)$.

\subsection{A Baseline Training Algorithm}

Observing that the feasible set of the dual problem in \eqref{eq:L2dual} is the standard simplex $\Delta^{n-1} = \bigl\{ \vec{\mu} \in \mathbb{R}^n \bigm| \vec{\mu} \succeq \vec{0}  \wedge \ipt{\vec{1}}{\vec{\mu}} = 1 \bigr\}$
and letting $\mat{H} \equiv \mat{K} \odot \opt{\vec{y}}{\vec{y}} + \opt{\vec{y}}{\vec{y}} + \tfrac{1}{C} \, \mat{I}$ and $f(\vec{\mu}) \equiv \tfrac{1}{2} \, \trn{\vec{\mu}} \mat{H} \, \vec{\mu}$ for brevity, \eqref{eq:L2dual} can be written more succinctly as
\begin{equation}
\label{eq:L2dualshort}
\amin{\vec{\mu} \in \Delta^{n-1}} \, f(\vec{\mu})
\end{equation}

This is now easily recognizable as a quadratic minimization problem over a compact convex set and therefore as a problem that can be solved using the Frank-Wolfe algorithm \cite{Frank1956-AAF}. In our practical examples below, we will thus consider iterative Frank-Wolfe optimization as baseline for $L_2$ SVM training.

\section{Gradient Flows for $L_2$ SVM Training}
\label{sec:flows}

Next, we devise a gradient flow for the optimization problem in \eqref{eq:L2dualshort} whose solution represents a parameter space trajectory that approaches an optimal choice of Lagrange multipliers from which one can compute SVM weight and bias parameters.

Since the term \emph{gradient flow} is nowadays sometimes re-appropriated to mean the backward flow of gradient information during neural network training, we emphasize that we understand it in its classical sense, namely
 
\begin{defn}
\label{def1}
Given a finite dimensional Euclidean vector space $\mathbb{R}^n$ and a smooth function $ f : \mathbb{R}^n \rightarrow \mathbb{R}$, a \emph{gradient flow} is a smooth curve $\vec{\mu}  : \mathbb{R} \rightarrow \mathbb{R}^n$, $t \mapsto \vec{\mu}(t)$ such that $\dot{\vec{\mu}}(t) = - \nabla f \bigl( \vec{\mu}(t) \bigr)$.
\end{defn}

As dynamical systems go, the behavior of finite dimensional gradient flows is rather simple. They either converge to a stationary point $\vec{\mu}_*$ of $f$ where $\nabla f(\vec{\mu}_*) = \vec{0}$, or they diverge as $t \rightarrow \infty$ \cite{Zinsl2015-SOE}. One can further show that, if $\vec{\mu}_*$ is an isolated stationary point and a local minimum of $f$, then $\vec{\mu}_*$ is an asymptotically stable equilibrium of $\dot{\vec{\mu}}(t) = - \nabla f \bigl( \vec{\mu}(t) \bigr)$ \cite{Absil2006-OTS,Helmke1994-OAD}.

\begin{figure*}[t!]
\subfigure[training data $\vec{x}_i \in \mathbb{R}^2$]{\includegraphics[height=3.cm]{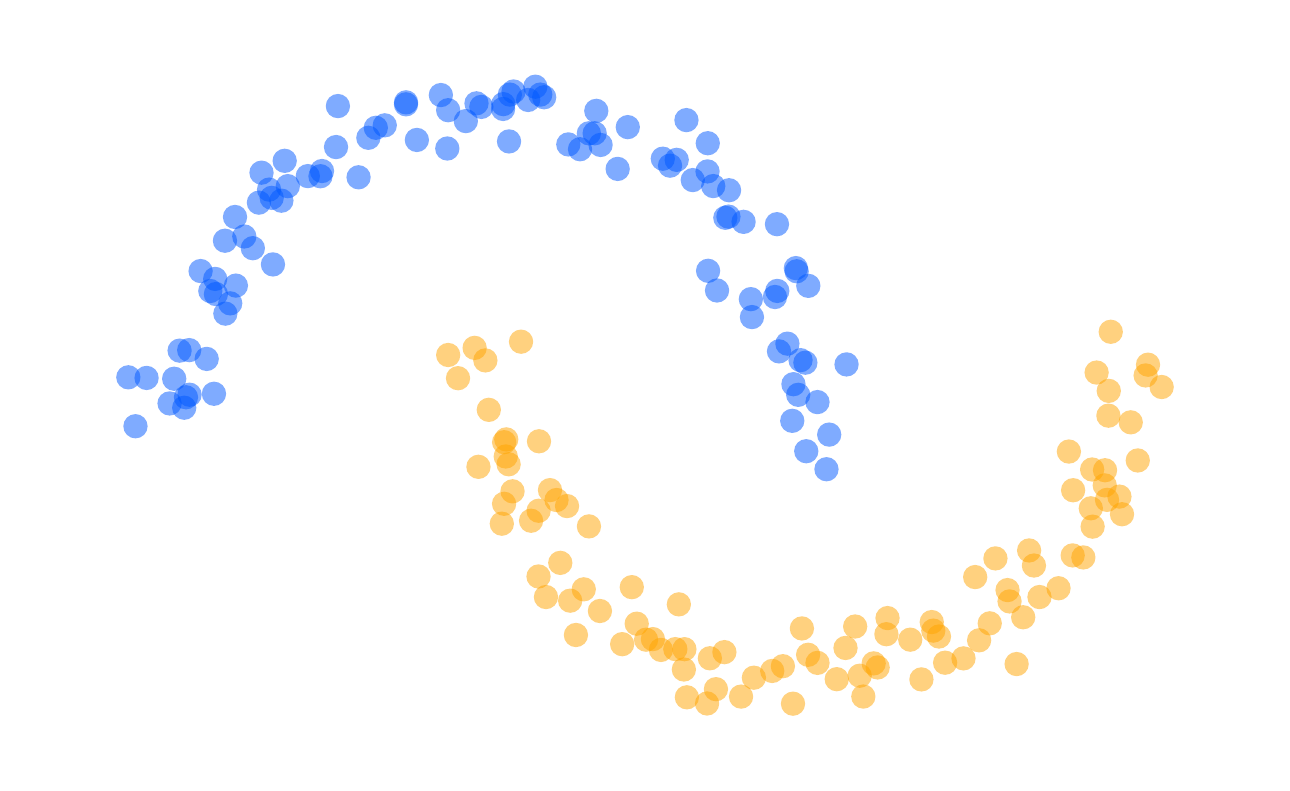}} \hfill
\subfigure[discrete Frank-Wolfe iterates $\vec{\mu}_t$]{\includegraphics[height=3.cm]{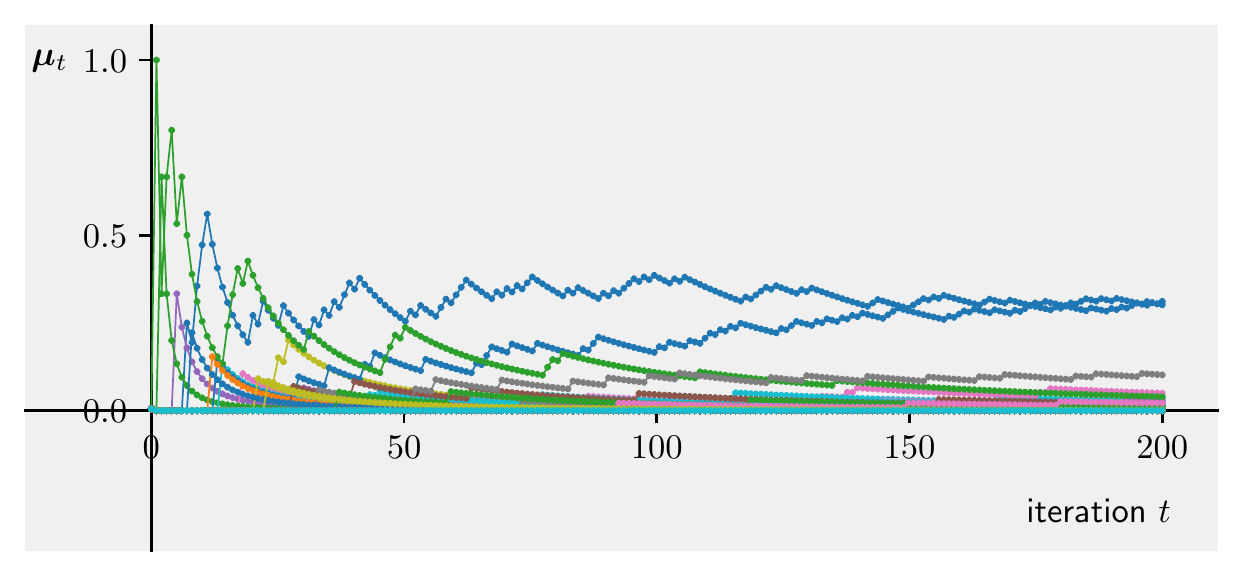}} \;
\subfigure[resulting classifier]{\includegraphics[height=3.cm]{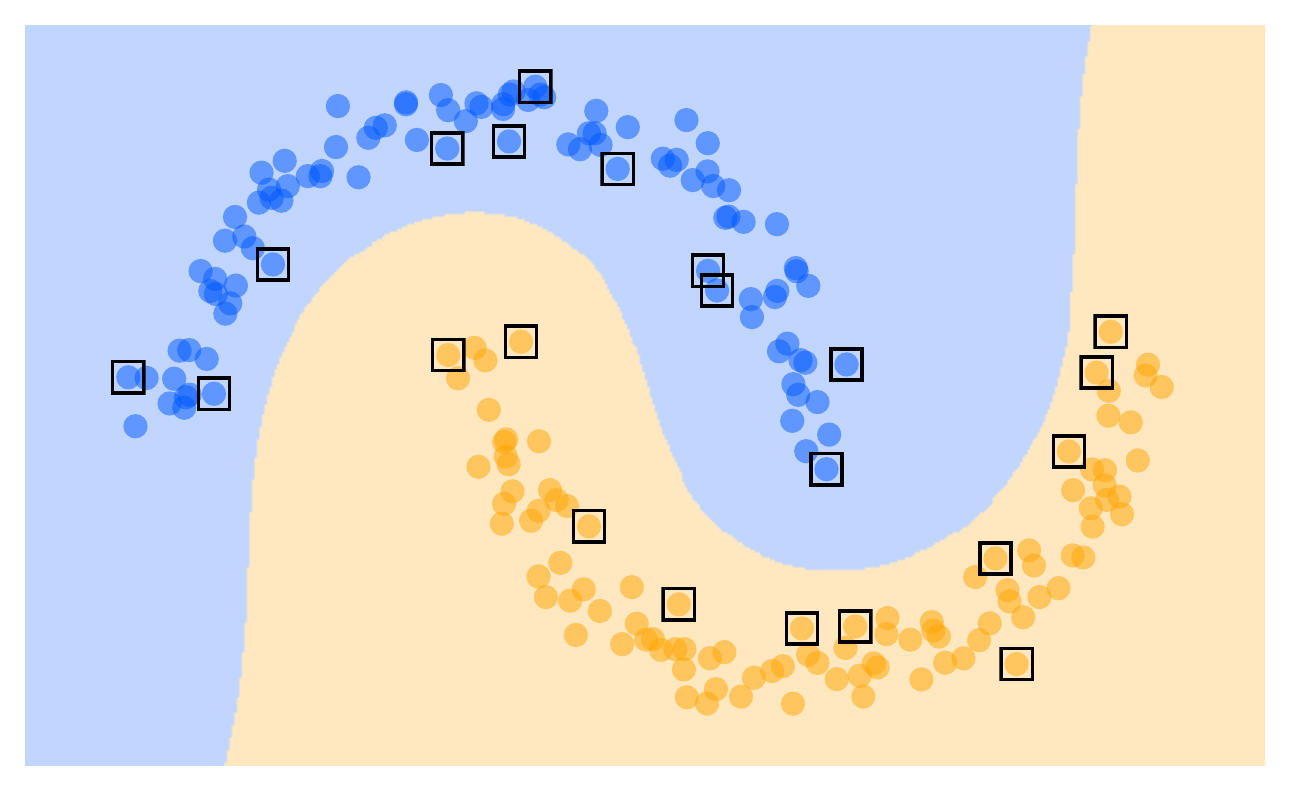}} 

\hfill
\subfigure[continuous time gradient flow $\vec{\mu}(t)$]{\includegraphics[height=3.cm]{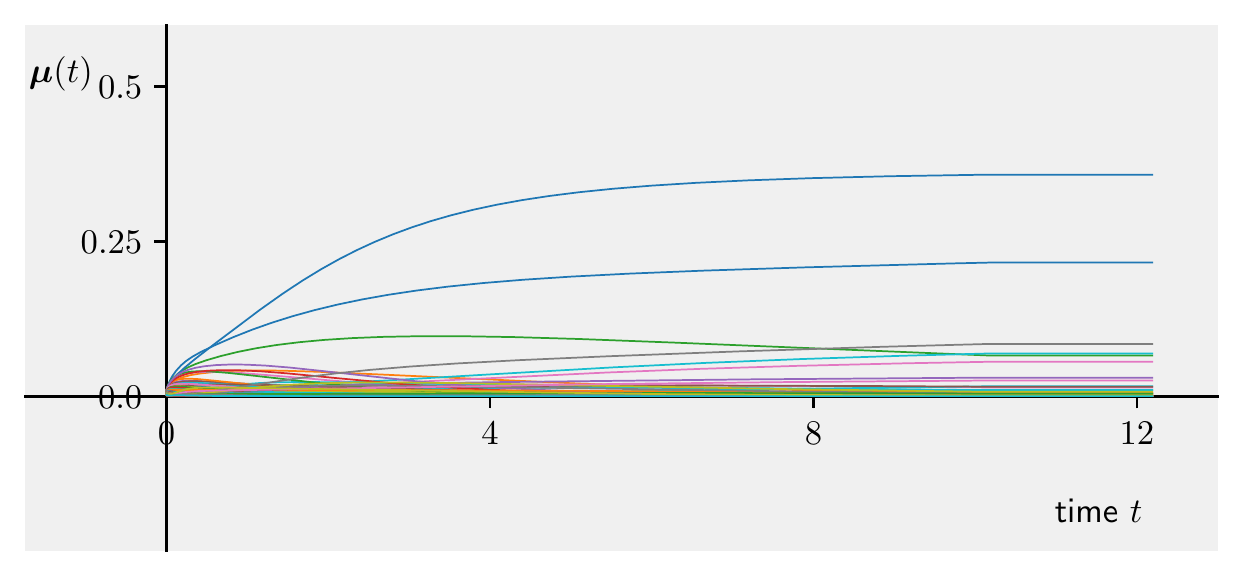}} \;
\subfigure[resulting classifier]{\includegraphics[height=3.cm]{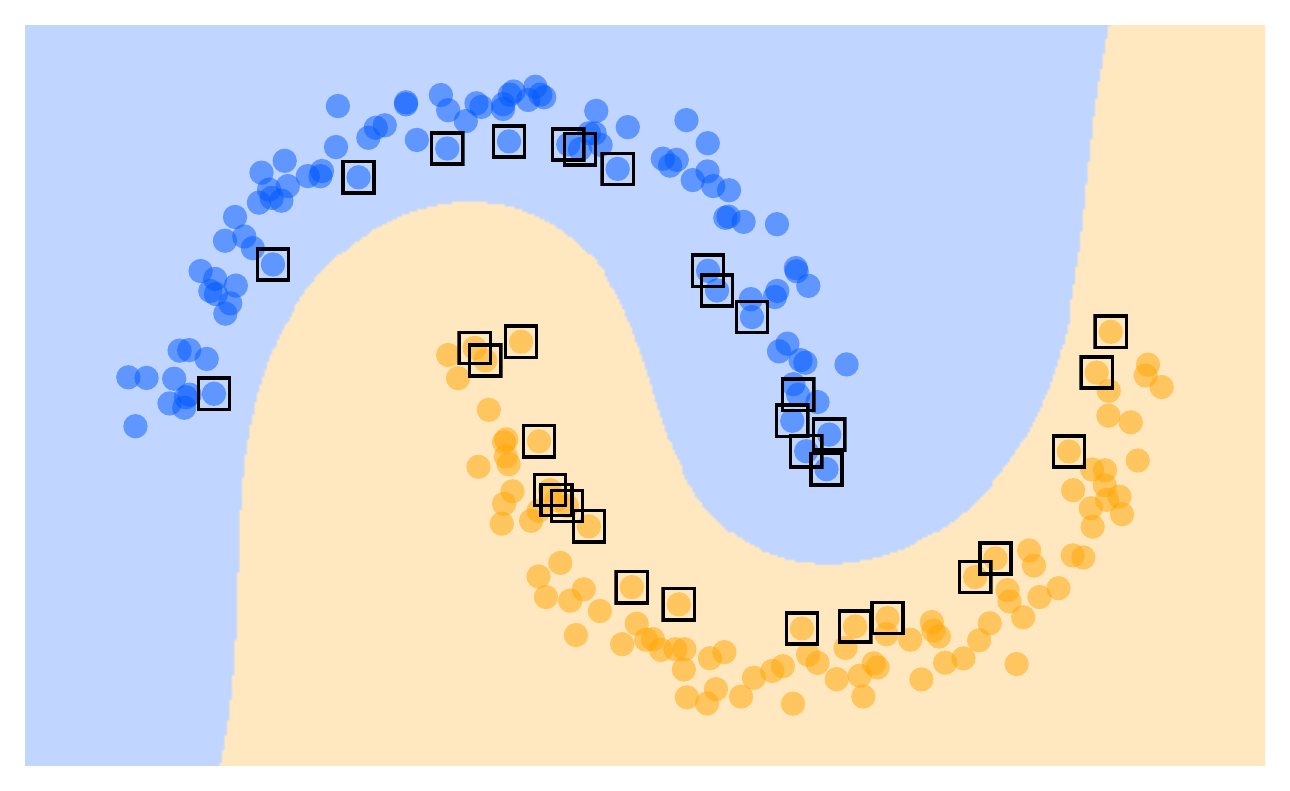}}
\caption{\label{fig:xmpl} Didactic examples of training kernel $L_2$ SVM classifiers via Frank-Wolfe optimization and via solving the flow in \eqref{eq:SVMflow}. The kernel function in both cases is a third order polynomial. The resulting classifiers are virtually indistinguishable; yet, the (feature space) separating hyperplane found via Frank-Wolfe iterations is supported by fewer support vectors than the one found from solving \eqref{eq:SVMflow}.}
\end{figure*}

Dropping the dependence on time $t$ for readability, the gradient of the objective function in \eqref{eq:L2dualshort} is given by $\nabla f ( \vec{\mu}) = \mat{H} \vec{\mu}$. However, a crucial difference between the gradient flow in the above definition and the optimization problem in \eqref{eq:L2dualshort} is that the former consider flows in $\mathbb{R}^n$ whereas solutions to the latter are confined to the standard simplex $\Delta^{n-1}$. We therefore note that $\Delta^{n-1}$ is an instance of a convex polytope $\set{P} = \bigl\{ \vec{\mu} \in \mathbb{R}^n \bigm| \vec{\mu} \succeq \vec{0}, \mat{A} \vec{\vec{\mu}} = \vec{b} \bigr\}$ and resort to 
a result by Helmke and Moore \yrcite{Helmke1994-OAD}. They show that an open convex polytope $\mathring{\set{P}} = \bigl\{ \vec{\mu} \in \mathbb{R}^n \bigm| \vec{\mu} \succ \vec{0}, \mat{A} \vec{\vec{\mu}} = \vec{b} \bigr\}$ can be endowed with a Riemannian metric and that the gradient flow $\dot{\vec{\mu}} = - \operatorname{grad} f ( \vec{\mu} )$ of $f : \mathring{\set{P}} \rightarrow \mathbb{R}$ with respect to this metric is given by
$\dot{\vec{\mu}} = -\bigl[ \mat{I} - \mat{D} \trn{\mat{A}} [ \mat{A} \mat{D} \trn{\mat{A}}]^{-1} \mat{A} \bigr] \mat{D} \nabla f ( \vec{\mu} )$ where $\mat{D} = \operatorname*{diag}(\vec{\mu})$ is an $n \times n$ diagonal matrix.

For the standard simplex, we have $\mat{A} = \trn{\vec{1}}$. Regarding the problem in \eqref{eq:L2dualshort}, we note that $\mat{D} \vec{1} = \vec{\mu}$ as well as $\ipt{\vec{1}}{\vec{\mu}} = 1$ so that $\trn{\vec{1}} \mat{D} \vec{1} = 1$. Restricted to the open simplex $\mathring{\Delta}^{n-1}$ and using $\nabla f ( \vec{\mu} ) = \mat{H} \vec{\mu}$, a gradient flow for \eqref{eq:L2dualshort} is therefore given by
\begin{align}
\label{eq:SVMflow}
\dot{\vec{\mu}} & 
= -\bigl[ \mat{I} - \mat{D} \vec{1} [ \trn{\vec{1}} \mat{D} \vec{1}]^{-1} \trn{\vec{1}} \bigr] \mat{D} \mat{H} \vec{\mu}
= -\bigl[ \mat{I} - \opt{\vec{\mu}}{\vec{1}} \bigr] \mat{D} \mat{H} \vec{\mu}
= -\mat{D} \mat{H} \vec{\mu} + \opt{\vec{\mu}}{\vec{\mu}} \mat{H} \vec{\mu}
%= -\bigl[ \mat{H} \vec{\mu} - \trn{\vec{\mu}} \mat{H} \vec{\mu} \vec{1} \bigr] \odot \vec{\mu}
\end{align}

Using software for numerical integration, this ordinary dynamical system can be practically solved for $\vec{\mu}(t)$. Figure~\ref{fig:xmpl} presents an example for how corresponding solutions behave in practice. 

Given the didactic, two-dimensional ``two moons'' data in Fig.~\ref{fig:xmpl}(a), we considered a third degree polynomial kernel $k(\vec{x}_i, \vec{x}_j) = (\ipt{\vec{x}_i}{\vec{x}_j} + 1)^3$ to set up matrix $\mat{H}$. Providing a baseline result, the plot in Fig.~\ref{fig:xmpl}(b) shows the evolution of discrete time Frank-Wolfe iterates $\vec{\mu}_t$. The plot in Fig.~\ref{fig:xmpl}(d) shows the evolution of the continuous time function $\vec{\mu}(t)$ obtained from solving \eqref{eq:SVMflow}. In both cases, the initial values for $\vec{\mu}$ were set to $\tfrac{1}{n} \vec{1}$. 

For the Frank-Wolfe iterates, we observe the typical jittering behavior which occurs when running plain vanilla versions of the algorithm; for the gradient flow solution, we observe a smooth evolution that quickly converges to a stationary point. Figures~\ref{fig:xmpl}(c) and (e) show the decision boundaries and support vectors of the classifier resulting from using $\vec{\mu}_{t=200}$ and $\vec{\mu}(t=12)$, respectively. The latter was rounded to five decimal places and renormalized because the flow in \eqref{eq:SVMflow} evolves in the open simplex $\mathring{\Delta}^{n-1}$ so that no entry of $\vec{\mu}(t)$ ever truly drops to zero. The decision boundaries of both $L_2$ SVMs are virtually indistinguishable; however, we note that the Frank-Wolfe solution for $\vec{\mu}$ is sparser than the gradient flow solution. The decision boundary resulting from the former is therefore computed from fewer support vectors than the decision boundary resulting from the latter.

Results and practical performance illustrated in this example are typical. That is, in ongoing work, we also observe the above behavior of gradient flows for SVM training when working with other data sets and different types of kernels. An interesting empirical observation in this regard is that polynomial kernels (including linear kernels where appropriate) always entail rapidly converging gradient flows whereas the popular Gaussian kernels seem to cause rather slow convergence to an equilibrium point. A mathematical analysis of this phenomenon is under way and results shall be reported soon.

\section{Outlook}
\label{sec:conclusion}

The initial results reported in this paper are part of ongoing research in which we investigate what kind of (classical) machine learning techniques lend themselves to implementations on energy efficient platforms or low power and low cost hardware. The overall goal behind these long-term efforts is to develop methods for robust on-sensor data analysis in environmental and agricultural application scenarios which address sustainability challenges in food production \cite{Bauckhage2012-ATM}. In this context, continuous time gradient flow formulations of machine learning objectives appear to be auspicious because analog circuits can solve differential equations while only requiring Milliwatts of energy \cite{Ulmann2013-AC}. 

Gradient flow formulations of learning tasks are not restricted to the ($L_2$) SVM classifiers considered in this paper. Corresponding expressions for least squares classifiers or linear discriminant classifiers are easy to come by and can also be devised for regression and forecasting problems \cite{Bauckhage2022-GFF}. The classic textbook by Helmke and Moore \yrcite{Helmke1994-OAD} provides a detailed account of the underlying theory and may serve as a source of inspiration for further developments. With respect to practical applications on edge devices, the challenges are therefore first and foremost technical rather than theoretical. In line with increased engineering efforts towards designing analog circuitry for machine learning \cite{Schuman2017-ASO,Haensch2019-TNG}, we are currently testing transistor-less hardware implementations of these models.

In the introduction, we mentioned yet another motivation for machine learning based on gradient flow formulations, namely quantum computing. Given that working quantum computers have now become a technical reality, efforts on their use at different stages of the machine learning pipeline are noticeably increasing \cite{Wittek2014-QML,Biamonte2014-QML}. Indeed, there already are proposals for quantum support vector machines \cite{Rebentrost2014-QSVM} and prototype implementations on existing quantum gate computers have been successfully realized  \cite{Piatkowski2022-TBA}. In this context, it is therefore interesting to note that classical SVM training can be accomplished via solving continuous time gradient flows because recently proposed quantum algorithms for diffrerential equation solving   \cite{Srivastava2019-BAF,Lloyd2020-QAF,Liu2021-EQA,Zanger2021-QAF} suggest alternative approaches towards quantum SVMs. 

Finally, it may be interesting to consider that the temporal evolution of a quantum system $\ket{\psi(t)}$ is governed by the ($\hbar = 1$) Schr\"odinger equation $\tfrac{d}{dt} \ket{\psi(t)} = -i \hat{H} \ket{\psi(t)}$ where the Hamiltonian $\hat{H}$ is a Hermitian operator ($\hat{H}^\dagger \hat{H} = \hat{H} \hat{H}^\dagger = \hat{I}$). Looking at the Schr\"odinger equation, we note that it can be seen as a gradient flow w.r.t. the weighted expectation value $\tfrac{i}{2} \bra{\psi(t)} \hat{H} \ket{\psi(t)}$. We further note that the matrix $[\opt{\vec{\mu}}{\vec{\mu}} -\mat{D}] \mat{H}$ in \eqref{eq:SVMflow} is Hermitian, too. All of this hints at the possibility of yet another quantum SVM training algorithm based on Grover's search \cite{Grover1996-AFQ} and we intend to further explore this idea in future work.

\vfill

\section*{Acknowledgments}

The authors gratefully acknowledge support by the Competence Center for Machine Learning Rhine-Ruhr (ML2R) funded by the Federal Ministry of Education and Research of Germany (grant no. 01IS18038C).% and by the Deutsche Forschungsgemeinschaft (DFG, German Research Foundation) under Germany’s Excellence Strategy -- EXC 2070 -- 390732324.

\newpage

\bibliographystyle{icml2022}
\bibliography{literature}

\end{document}